\documentclass[conference]{IEEEtran}
\IEEEoverridecommandlockouts
\usepackage{cite}
\usepackage{amsmath,amssymb,amsfonts}
\usepackage{algorithmic}
\usepackage{graphicx}
\usepackage{textcomp}
\usepackage{xcolor}
\usepackage{multirow}
\def\BibTeX{{\rm B\kern-.05em{\sc i\kern-.025em b}\kern-.08em
    T\kern-.1667em\lower.7ex\hbox{E}\kern-.125emX}}
\begin{document}

\title{

MARL: Multimodal Attentional Representation Learning for 
Disease Prediction
}

\author{\IEEEauthorblockN{1\textsuperscript{st} Ali Hamdi}
\IEEEauthorblockA{\textit{RMIT University} \\
Melbourne, Australia \\
ali.ali@rmit.edu.au}
\and
\IEEEauthorblockN{2\textsuperscript{nd} Amr Aboeleneen}
\IEEEauthorblockA{\textit{Qatar University} \\
Doha, Qatar \\
aa1405465@qu.edu.qa}
\and
\IEEEauthorblockN{3\textsuperscript{rd} Khaled Shaban}
\IEEEauthorblockA{\textit{Qatar University} \\
Doha, Qatar \\
khaled.shaban@qu.edu.qa}
}

\maketitle

\begin{abstract}
Existing learning models often utilise CT-scan images to predict lung diseases. These models are posed by high uncertainties that affect lung segmentation and visual feature learning. We introduce MARL, a novel \textit{M}ultimodal \textit{A}ttentional \textit{R}epresentation \textit{L}earning model architecture that learns useful features from multimodal data under uncertainty. We feed the proposed model with both the lung CT-scan images and their perspective historical patients' biological records collected over times. Such rich data offers to analyse both spatial and temporal aspects of the disease. MARL employs Fuzzy-based image spatial segmentation to overcome uncertainties in CT-scan images. We then utilise a pre-trained Convolutional Neural Network (CNN) to learn visual representation vectors from images. We augment patients' data with statistical features from the segmented images. We develop a Long Short-Term Memory (LSTM) network to represent the augmented data and learn sequential patterns of disease progressions. Finally, we inject both CNN and LSTM feature vectors to an attention layer to help focus on the best learning features. We evaluated MARL on regression of lung disease progression and status classification. MARL outperforms state-of-the-art CNN architectures, such as EfficientNet and DenseNet, and baseline prediction models. It achieves a $91\%$ $R^2$ score, which is higher than the other models by a range of $8\%$ to $27\%$. Also, MARL achieves $97\%$ and $92\%$ accuracy for binary and multi-class classification, respectively. MARL improves the accuracy of state-of-the-art CNN models with a range of $19\%$ to $57\%$. The results show that combining spatial and sequential temporal features produces better discriminative feature.
\end{abstract}


\section{Introduction}
Deep representation learning models are proposed to learn discriminative features in various applications. Recently, lung disease prediction tasks, be it progression regression or classification, have gained much attention due to the pandemic of COVID-19. Existing prediction models are challenged by uncertainty issues when determining the correct disease patterns \cite{cottin2014diagnosis,kafaja2018reliability,hamdi2021spatiotemporal}. These uncertainty issues effect lung disease prediction when performing lung segmentation and feature representation learning. 

Lung segmentation is challenging due to the fuzziness of the visual compositions of the CT-scan images. These images contain radio-density Hounsfield scores that represent other human body parts. Conventional methods often depend on tissue thresholds over the CT-scan Hounsfield. Such methods also employ various morphological operations such as dilation to cover the lung nodules at far borders. However, these methods suffer from the uncertainty when separating the lung tissues. Therefore, we propose to apply Fuzzy-based spatial segmentation over the lung CT-scans to reduce noisy spots and spurious blobs in the images \cite{tripathy2014image}. We then employ a pre-trained Convolutional Neural Network (CNN) to learn the visual features of the images.

Visual representation learning models depend, in most cases, on CNN to capture useful patterns in a given image \cite{tan2019efficientnet}. However, CNNs suffer from challenges such as limited local structural information as they are designed to learn local descriptors using receptive fields \cite{luo2016understanding}. This, in turn, leads to losing important global structures. We propose to address this issue by augmenting the CNN visual features with global temporal characteristics from patients biological and health records. We train our proposed hybrid model to learn these temporal features through a Long Short-Term Memory (LSTM) network. However, LSTM learns features from sequences at a fixed length limiting the significance of the learning feature space. We propose to overcome this limitation by employing an attention layer that learns what should be learnt from the visual CNN and sequential LSTM features. 

\begin{figure*}[ht]
\begin{center}
   \includegraphics[width=1\linewidth]{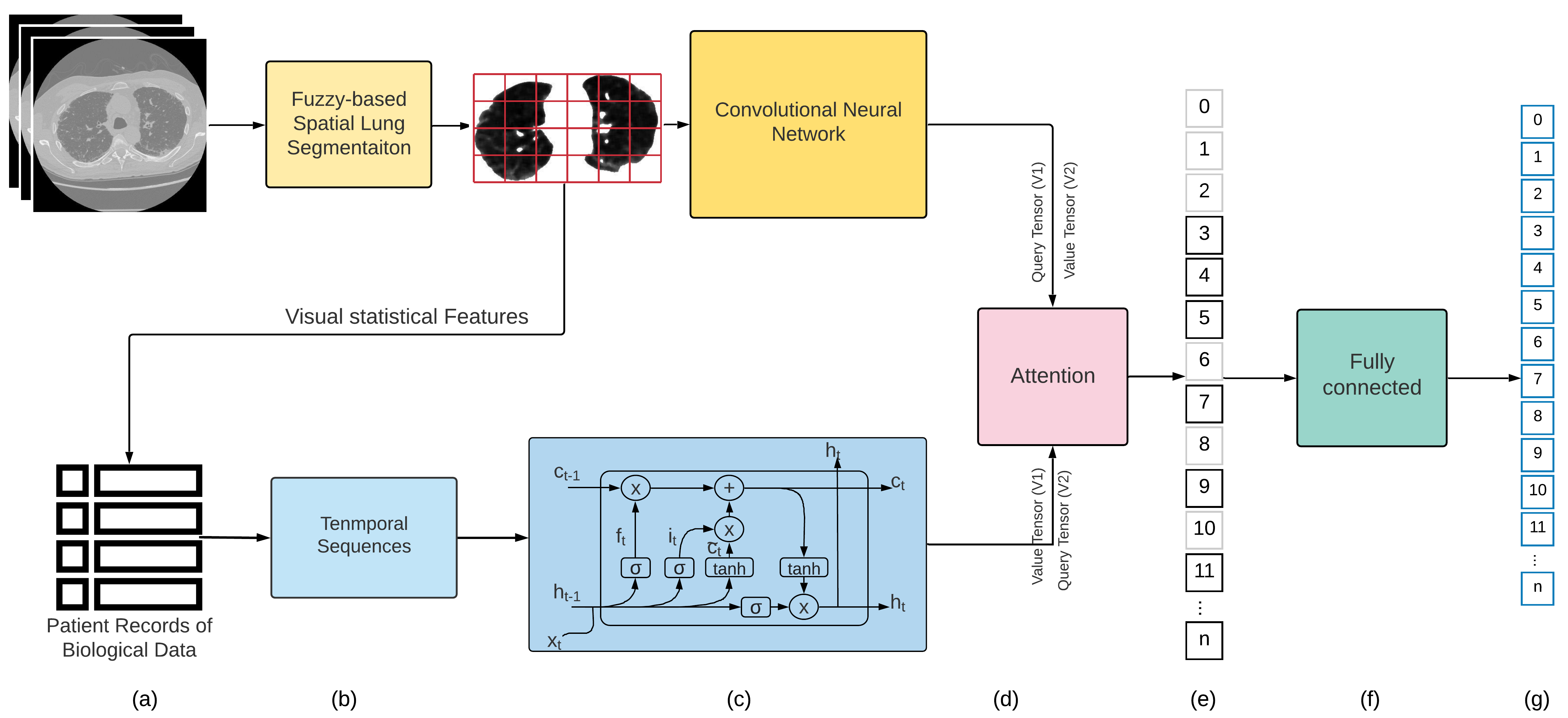}
\end{center}
   \caption{Workflow of the proposed Multimodal Attentional Representation Learning model architecture (MARL) for IPF lung disease prediction. (a) Inputing Lung CT-scan images and patient data. (b-top) Applying Fuzzy-based spatial lung segmentation. (b-bottom) Preparing temporal sequences of the patient attributes and extracting statistical features from the segmented lung images. (c) Learning visual features using pre-trained CNN and temporal features using LSTM. (d and e) Injecting the visual and sequential feature vectors to an attention layer that learns the best set of weights. (f and g) Adding a fully-connected layer to learn the final feature vector from the multimodal data.}
\label{framework}
\end{figure*}

As a case study, we utilise a public dataset for Idiopathic Pulmonary Fibrosis (IPF) lung disease \cite{IPF_kaggle_2020}. IPF scars lung tissues and it worsens over time for unknown causes \cite{kafaja2018reliability}. When infected, lungs cannot take in the required amount of oxygen due to difficulty of breathing. The dataset contains both CT-scan images and patients' biological and health records with different attributes collected over periods of times. Such multimodal data harnesses spatial visual features of the disease and temporal patient attributes. Our proposed MARL, a novel Multimodal Attentional Representation Learning model architecture, has multiple components as visualised in Fig. \ref{framework}. MARL starts with preprocessing the input of the lung CT-scan images using a fuzzy-based spatial segmentation and preparing the temporal sequences from their corespondent patients' data records. Two different deep learning networks, including CNN and LSTM, are then employed to learn the visual and temporal features, respectively. The encoded pairs of feature vectors are injected into an attention layer. The final feature vectors are propagated through a fully-connected layer. These final feature vectors are evaluated against multiple downstream tasks such as regressing the disease progression and classifying the disease status.

In summary, the multimodal learnt feature vectors, produced by MARL, offer to solve the uncertainty issues in lung disease prediction through the following contributions:
\begin{itemize}
    \item Producing accurate visual representation vectors by improving the CNN feature learning through a Fuzzy-based spatial segmentation.
    \item Developing effective temporal feature learning from the patient biological records and statistical information of the CT-scan images using an LSTM network.
    \item Introducing an attention mechanism that improves the feature representation by focusing on the critical input sequences and image parts.
    \item Carrying out an extensive experimental work to evaluate the proposed model on different lung disease prediction tasks such as lung declination regression and disease status classification.
\end{itemize}

The rest of the paper is organised as follows. Section \ref{rw} reviews related works and contrast that with our work in this paper. Section \ref{rm} explains the proposed MARL. Section \ref{exp} describes experimental setups, and discusses the results of performance evaluations. Section \ref{con} concludes the paper. 

\section{Related Work}\label{rw}
Recent research efforts have used CNNs visual representation learning to advance multiple applications \cite{zhang2019aet,kolesnikov2019revisiting,hamdi2020drotrack}. CNN-based methods are widely utilised to produce visual feature vectors from CT-scan images for disease prediction \cite{yamashita2018convolutional}. Lung images can be categorised based on the disease status \cite{raghu2011official,bueno2018updated}. A CNN network was developed by the authors of \cite{walsh2018deep} to classify the disease status into positive, possible-to-have, and negative. They collected a lung disease dataset of $1,157$ high-resolution images. Their experimental results showed the superiority of the deep learning models against the radiologists in both accuracy and speed. However, their results depended on a small dataset which limits the feature learning space. In this paper, we utilise a dataset of $33,026$ CT-scan lung images in addition to the patient tabular data. Besides, their dataset was annotated by one expert who might have erroneous decisions. On contrast, the dataset we use is created and published by Open Source Imaging Consortium that made substantial cooperative efforts between the academia and healthcare industry.

State-of-the-art of CNN-based models proposed various network architectures to improve image representation learning \cite{simonyan2014very,he2016deep,huang2016deep,he2016identity,huang2017densely,Sandler_2018_CVPR,howard2017mobilenets,zoph2018learning,chollet2017xception,tan2019efficientnet}. CNN-based models are designed to have large sets of layers to adapt to the increasing size and complexity of the training data \cite{hamdi2021flexgrid2vec}. However, they usually suffer from the problem of overfitting when having relatively small training data. Recently, the overfitting problem has been addressed by different techniques such as data augmentation \cite{masi2016we}. Moreover, CNNs neglect useful structures due to the limitations of their receptive fields and isotropic mechanism \cite{luo2016understanding}. Therefore, we propose to combine the visual CT-scan data with their correspondent patients' data. Such multimodality adds useful features that increase the accuracy of lung disease prediction. The patient data is a set of patient attributes and disease progression measurements over time. Therefore, we employ LSTM to learn temporal features combined with the CNN visual representations in order to improve prediction. LSTM has been recently used to predict different disease developments such as Alzheimer \cite{hong2019predicting}, hand-foot-mouth \cite{gu2019method}, and COVID-19 \cite{chimmula2020time}. LSTM learns better at equal intervals of regular spaced timestamps. Notwithstanding that CT-scan data are often collected at the patient needs making an irregular data collection sequences. The authors in \cite{gao2019distanced} utilised adapted LSTM to learn irregular temporal data points for lung cancer detection. Moreover, LSTM learns features from sequences at a fixed length which effects the learning of the feature space. Therefore, we design our proposed MARL to combine CNN and LSTM to overcome their limitations.

The hybridisation between CNN and LSTM tends to have a potential increase in disease prediction accuracy. Recent work in \cite{marentakis2021lung} has reported that a hybrid model of LSTM and CNN outperformed the human experts in lung disease classification. However, their work did not consider the uncertainty in segmentation, and they used a small dataset of $102$ patients. Therefore, there is still a need to address the above-discussed limitations and uncertainties in LSTM and CNN networks. We utilise Fuzzy-based spatial segmentation to improve the lung segmentation before the convolutional feature extraction. Using multimodal datasets contributes to accurate lung disease prediction \cite{tang2020elaboration,li2020deep}. The work in \cite{subramanian2020multimodal} predicted the recurrence of lung cancer based on a multimodal fusion of tomography images and genomics. However, using CNN and LSTM on multimodal data adds more complexity to the training process. We design an attentional neural layer at the bottleneck that connects the CNN and LSTM vectors with the fully-connected layer. This attention mechanism is designed to make the model focuses on essential features in the input sequences. The authors in \cite{qiao2019mnn} combined both medical codes and clinical text notes to implement multimodal attentional neural networks. Similarly, our proposed model, MARL, combines different data modalities such as CT-scan images, and patients' biological and health records. MARL, extract useful representations from CT-scan images, patient data, and visual statistical information. 


\section{MARL: Multimodal Attentional Representation Learning}\label{rm}
We propose a novel representation learning model architecture for lung disease prediction. The proposed model is designed to address uncertainty issues that affect the downstream prediction tasks such as disease progression regression and disease status binary and multi-class classification. In this section, we explain the workflow phases of MARL.

\subsection{Preprocessing}
The given CT-scan images in the dataset have multiple issues regarding colour exposure and varying sizes. We start by correcting the black exposure to ensure high quality of the subsequent feature extraction step. We also crop and scale-up the dataset images to match a unified size. 

Around the lung, the CT-scan images include other human body parts, such as bones and blood vessels. Therefore, the images must be segmented to extract the lung parts only. Pixels in the CT-scan images contain radio-density scores. The pixel value represents the mean attenuation of the tissue scale from $-1,024$ to $+3,071$ at Hounsfield scale. Hounsfield unit (HU) is a scaled linear transformation of the radio-density's attenuation coefficient measurements. HU values are calculated as in Eq. \ref{hu}.
\begin{equation}\label{hu}
    hu = pixel_{value} * slope + intercept
\end{equation}
where slope and intercept are stored in the CT-scan file. The projected HUs are interpreted according to the ranges, such as, bone from $+700$ to $+3000$ and lung to be $-500$. However, segmenting the lung based on these numbers is cumbersome. The visual composition of the different body parts is uncertain at multiple locations. Therefore, we implement spatial segmentation based on Fuzzy C-Means (FCM) applied on the HUs.

\subsection{Lung Segmentation with Spatial Fuzzy C-Means}
The lung segmentation suffers from uncertainty due to the fuzzy area around the lung. This fuzziness happens because of the nature of the HU values that represent various human body parts around the lung. Using the Fuzzy spatial C-means has advantages over the classical Fuzzy C-Means. The latter is sensitive to noisy parts in the given images. Besides, the classic FCM expects data to have robust, and separated partitions to implement useful membership functions. Nonetheless, in our case, this assumption is not valid due to the high-dependency among the image segments. Spatial FCM computes the likelihood of a neighbourhood-pixel belongs to a specific segment, e.g., lung. The Fuzzy member function uses the spatial likeliness score to calculate the membership value. The work in \cite{tripathy2014image} proposed to compute the membership values ($m$) based on score of spatial similarity and degree of hesitation, as in Eq. \ref{msh}.
\begin{equation}\label{msh}
m_{ij}=\frac{u_{ij}^{p}h_{ij}^{q}}{\sum\limits_{k=1}^{c}u_{kj}^{ p}h_{kj}^{q}}
\end{equation}
where $m_{ij}$ denotes membership values of a neighbourhood pixel with coordinates of $i$ and $j$, $u$ denotes membership function calculated based on the degree of hesitation score, and $p$ regulates the initial membership's weights, $q$ controls spatial functions, and $h$ is the spatial function. We at that stage apply standard morphological transformation methods. Specifically, we utilise erosion and dilation in order to remove the noise remain thereafter segmentation. 


\subsection{Patient Biological Data Enrichment}
The dataset contains patient records of biological information. These tabular data are temporally tagged with different timestamps of their collections. The dataset has the following columns:
\begin{itemize}
    \item Patient unique identifiers that link the biological data with the CT-scan images.
    \item Week numbers of which the CT-scan had been taken.
    \item FVC, the Forced Vital Capacity recorded for the patient lung in that week.
    \item The percent: a calculated field that estimates the FVC of a patient as a percentage of the average FVC for patients with similar attributes.
    \item Age of the patient.
    \item Sex denoting the gender of each patient.
    \item Smoking Status of a patient as smoker, non-smoker, or ex-smoker.
\end{itemize}

\begin{figure}\label{dist}
    \centering
    \includegraphics[width=0.4\textwidth]{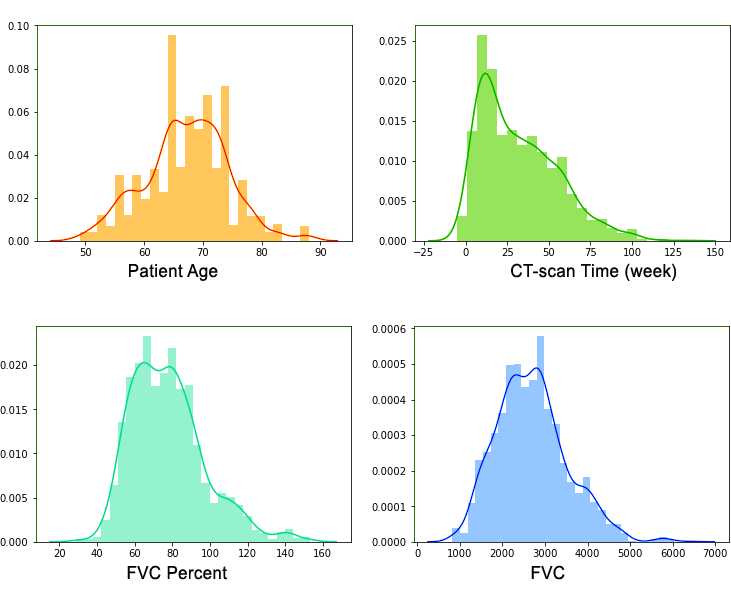}
    \caption{Unbalanced distributions of the biological data.}
    \label{fig:imbalanced_nd}
\end{figure}

The biological dataset is unbalanced, as shown in Fig. \ref{dist}. The patient ages, CT-scan times, and FVC percentages and scores show unbalanced distributions. Moreover, the dataset has records of males more than females, ex-smokers more than smokers and non-smokers. This fact adds more uncertainty due to the bias towards particular categories. Therefore, we enrich the biological tabular data with some visual statistical features. We compute the kurtosis, volume, mean, skewness, and moments for each CT-scan segmented lung. Adding such visual statistical information to the tabular biological data tends to be useful for achieving high accuracy. Moreover, we implement an LSTM network to overcome the uncertainty issue due to the data unbalance by leaning useful sequential temporal patterns. 

\subsection{Multimodal Representation Learning}
We implement two deep neural networks. First, Convolutional Neural Network to extract the CT-scan images' visual features. Second, a double Long Short Term Memory Network to learn sequential temporal features from the biological and visual statistical data. 

\subsubsection{Convolutional Neural Networks}
We employ Efficient-Net \cite{tan2019efficientnet}, which has recently outperformed other pre-trained networks in accuracy, size, and efficiency. 

A CNN network is typically composed of one or more convolutional layers. Each layer $i$ represents a function as in Eq. \ref{cnnl}.
\begin{equation}\label{cnnl}
    Y_i = \mathcal{F}_i(X_i)
\end{equation}
where $Y_i$ is output, $\mathcal{F}_i$ is a convolution operator, and $X_i$ is the input features. $X_i$ is a tensor of the shape of $<H_i, W_i, C_i>$ with $H_i$, $W_i$, and $C_i$ denote the tensor spatial dimension and channel dimension. Thus, a convolutional network can be composed of multiple stages or groups of CNN layers as in Eq \ref{cnn} \cite{he2016deep}.
\begin{equation}\label{cnn}
    \mathcal{N} = \bigoplus_{i=1 \ldots s} \mathcal{F}_{i}^{L_{i}}\left(X_{\left\langle H_{i}, W_{i}, C_{i}\right\rangle}\right)
\end{equation}
where $\mathcal{N}$ represents the CNN neural network that has $\mathcal{F_i}$ layer repeated $L_{i}$ times in stages $i$. Most of the CNN architectures aim to find the best layer design and network scale of length $L_i$ and width $C_i$. The employed Efficient-Net is designed to maximise the network performance according to the available resources. 
\begin{equation}
    \begin{array}{ll}
    \max _{d, w, r} \text { Accuracy }(\mathcal{N}(d, w, r)) \\
    \text { s.t. }  \mathcal{N}(d, w, r) = \bigoplus_{i=1 \ldots s} \hat{\mathcal{F}}_{i}^{d \cdot \hat{L}_{i}}\left(X_{\left\langle r \cdot \hat{H}_{i}, r \cdot \hat{W}_{i}, w \cdot \hat{C}_{i}\right\rangle}\right) \\
    \text{Memory}(\mathcal{N}) \leq \text { target\_memory } \\
    \text{FLOPS}(\mathcal{N}) \leq \text { target\_flops }
    \end{array}
\end{equation}
where $w,d,r$ denote the width, depth, and resolution coefficients for scaling the network, $\hat{\mathcal{F}}_{i}, \hat{L}_{i}, \hat{H}_{i}, \hat{W}_{i}, \hat{C}_{i}$ are the predefined network parameters. The network depth scaling is a popular task in CNN. Most recent CNN networks assume that deeper networks capture rich and complex features. However, this intuition is difficult to train because of the vanishing gradient problem \cite{zagoruyko2016wide}. Recent advances have proposed to alleviate this issue via batch normalisation \cite{ioffe2015batch} and skip connection \cite{he2016deep}. However, the network performance diminishes, and the accuracy does not increase even if the network depth is increased \cite{tan2019efficientnet}. Wider networks are also assumed to be able to capture more fine-grained features and can also be trained easily \cite{tan2019mnasnet}. However, having a wide but shallow network suffers difficulty learning high-level representations. Training a CNN on high-resolution images tends to produce better representations. In our case, the CT-scan images are in different resolutions. This varying image size also adds to the uncertainty problem in capturing useful visual representations. Therefore, we augment the visual feature vector with another feature vector that can be learnt from the biological and visual statistical data through an LSTM model.

\subsubsection{Long Short Term Memory}
The patient temporally recorded biological data are combined with visual statistical features from the CT-scan images. This data are then padded into identical sequences to be ready for LSTM learning. LSTM is a type of Recurrent Neural Networks. LSTM is featured by having feedback connections by which it controls a sequence of data inputs instead of single inputs. An LSTM can be implemented with a forget gate as Eq. \ref{lstm}
\begin{equation}\label{lstm}
\begin{aligned}
f_{t} &=\sigma_{g}\left(W_{f} x_{t}+U_{f} h_{t-1}+b_{f}\right) \\
i_{t} &=\sigma_{g}\left(W_{i} x_{t}+U_{i} h_{t-1}+b_{i}\right) \\
o_{t} &=\sigma_{g}\left(W_{o} x_{t}+U_{o} h_{t-1}+b_{o}\right) \\
\tilde{c}_{t} &=\sigma_{c}\left(W_{c} x_{t}+U_{c} h_{t-1}+b_{c}\right) \\
c_{t} &=f_{t} \circ c_{t-1}+i_{t} \circ \tilde{c}_{t} \\
h_{t} &=o_{t} \circ \sigma_{h}\left(c_{t}\right)
\end{aligned}
\end{equation}
where 
$f_{t} \in \mathbb{R}^{h}$ denotes the activation vector of the LSTM forget gate,
$x_{t} \in \mathbb{R}^{d}$ represents the input vector to the utlised LSTM network, 
$i_{t} \in \mathbb{R}^{h}$ is the activation vector of the LSTM input and update gate,
$o_{t} \in \mathbb{R}^{h}$ is the activation vector of the LSTM output gate, 
$\tilde{c}_{t} \in \mathbb{R}^{h}$ represents the activation vector of the LSTM cell input, 
$c_{t} \in \mathbb{R}^{h}$ denotes the cell state vector, 
$h_{t} \in \mathbb{R}^{h}$ denotes the hidden state or output vector of the LSTM unit,
$W \in \mathbb{R}^{h \times d}$ denotes the weights of the input, 
$U \in \mathbb{R}^{h \times h}$ denotes the weights of the recurrent connections, and 
$b \in \mathbb{R}^{h}$ denotes the parameters of the bias vector learnt throughout the training process. The superscripts $h$ and $d$ denote the number of hidden units and input features, respectively. We implement two LSTM layers on the biological and visual statistical data. The output feature vector will be injected into an attention layer alongside the previous CNN visual feature vector as denoted in Fig. \ref{framework}. 

\subsubsection{Attention Layer and Feature Vector Concatenation}
At this stage, we have extracted two feature vectors from the CNN and LSTM models. We then pass these feature vectors to an attention layer to learn the best features. We implement a dot-product attention layer based on Luong attention \cite{luong2015effective}. The attention layer expects query $T_q$ and value $T_v$ tensors. It starts with calculating the scores of the dot product operations as $scores = T_q * T_v$ computing the query-value dot product. Then, the $scores$ are used to compute the distribution based on softmax function as in Eq. \ref{softmax}. 
\begin{equation}\label{softmax}
    \operatorname{Softmax}\left(x_{i}\right)=\frac{\exp \left(x_{i}\right)}{\sum_{j} \exp \left(x_{j}\right)}
\end{equation}
The output distribution vector is utilised to create a linear combination of the value tensor $T_v$. The output attention vector is passed to a fully-connected layer to learn the final representation vector.

\section{Experimental Work}\label{exp}
We present a set of experiments to highlight MARL's efficacy. We implement its components as follows:
\begin{itemize}
    \item Preprocessing the CT-scan images 
    \begin{itemize}
        \item Correcting the black exposure. 
        \item Unifying the image sizes. 
        \item Using the Fuzzy spatial C-Means to segment the lung.
    \end{itemize}
    \item Preprocessing the patient's health records, as follow:
    \begin{itemize}
        \item Adding visual statistical features of the correspondent images.
        \item Making identical sequences to be ready for the LSTM sequential learning.
    \end{itemize}
    \item Utilising multiple state-of-the-art CNN architectures to learn the visual features vectors form the lung images.
    \item Using a double LSTM architecture to learn sequential temporal features from the health and visual statistical records. 
    \item Implementing two versions of the attention mechanism, as follows:
    \begin{itemize}
        \item MARL V1: Using the CNN visual feature vectors as query tensors to find the best features for the LSTM to learn. 
        \item MARL V2: Using the LSTM feature vectors as query tensors to learn where to focus in the images.
    \end{itemize} 
    \item Adding a fully-connected layer to learn the final feature vectors.
\end{itemize}
The learnt feature vectors at the last step are now ready to be consumed by any downstream task. We introduce lung disease tasks as follows:
\begin{itemize}
    \item Estimating disease progression by adding a regression layer on top of MARL.
    \item Classifying the lung disease status on binary and multi-class models.
\end{itemize}
In the next subsection, we discuss the utilised dataset and experimental results of both the regression and classification tasks.

\subsection{IPF Lung Disease Dataset} 
\begin{figure}
    \centering
    \includegraphics[width=0.4\textwidth]{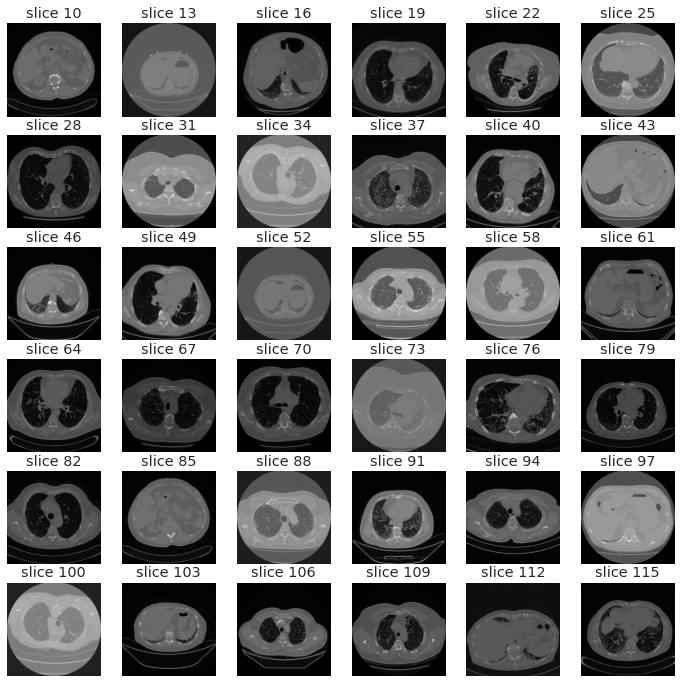}
    \caption{Sample CT slices}
    \label{fig:sam_slices}
\end{figure}

The data includes $1,549$ patients' health records, $33,026$ CT-scan images, $880$ of them are used for testing. A sample of CT slices are shown in Fig. \ref{fig:sam_slices}. Some CT-scan images have different resolutions, and some need colour correction. Moreover, the dataset provides unbalanced data categories where the number of males exceeds the number of females, and the number of ex-smokers exceeds the smokers and non-smokers. 


\subsection{Regression of IPF Lung disease progression}
\begin{figure}
    \centering
    \includegraphics[width=0.3\textwidth]{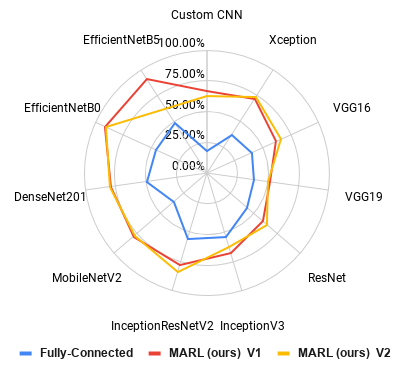}
    \caption{A radar-chart shows the performances of CNN networks and the proposed MARL model.}
    \label{Radar-MARL-Reg-CNN}
\end{figure}

\begin{figure}
    \centering
\includegraphics[width=0.3\textwidth]{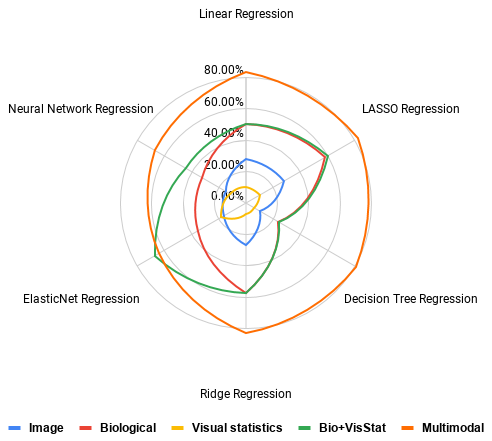}
    \caption{A radar-chart shows the performances of the regression models using different data modalities.}
    \label{radar-MARL-reg-multimodal}
\end{figure}

Table \ref{reg} reports the experimental results of lung disease regression using state-of-the-art CNNs on CT-scan images with three different setups, i.e., MARL V1, V2, and a fully-connected regression layers on top of the CNN architectures to extract the visual features as explained earlier.  
The results show that our model outperforms the other models. MARL manages to improve the accuracy of all CNN networks. The performance improvements range from $11\%$ to $49\%$ as shown in Table \ref{reg}. Fig. \ref{Radar-MARL-Reg-CNN} shows a comparison between the three experimental setups using a radar chart and the superiority of MARL V1 and V1 is noticeable. We also evaluate the impact of each data modality and their combinations. Table \ref{reg_mm} and Fig. \ref{radar-MARL-reg-multimodal} compare the performance of the regression models using each data source separately and combined. Using the biological data alone produces better results than the image and visual statistical data individually. The multimodal dominates the results over the other regression models. Besides, the table lists the performance results of MARL V1 and V2 on the multimodal data. MARL outperforms all with $91\%$ and $89.6\%$ $R^2$ scores for V1 and V2 setups, respectively. These results are higher than the other regression models by a range of $8\%$ to $27\%$ $R^2$ when compared to the results of the other models that are provided with multimodal data.

\begin{table}[]
\centering
\caption{Regression results of IPF lung disease progression using various CNN models and MARL V1 \& V2.}\label{reg}
\begin{tabular}{|l|c|c|c|}
\hline
\multicolumn{1}{|c|}{Model} & \multicolumn{1}{l|}{Fully-Connected} & \multicolumn{1}{l|}{\textbf{MARL V1}} & \multicolumn{1}{l|}{MARL V2} \\ \hline
VGG19 \cite{simonyan2014very} & 39\% & 52.10\% & 50\% \\ \hline
Custom CNN & 18\% & 67\% & 63\% \\ \hline
InceptionV3 \cite{szegedy2016rethinking} & 54\% & 68\% & 63\% \\ \hline
ResNet \cite{he2016deep} & 43\% & 60\% & 65\% \\ \hline
VGG16 \cite{simonyan2014very} & 40\% & 62\% & 66\% \\ \hline
Xception \cite{chollet2017xception} & 37\% & 72\% & 74\% \\ \hline
MobileNetV2 \cite{Sandler_2018_CVPR} & 36\% & 79\% & 78\% \\ \hline
DenseNet201 \cite{huang2017densely} & 49\% & 79\% & 80\% \\ \hline
InceptionResNetV2 \cite{szegedy2017inception} & 56\% & 78\% & 84\% \\ \hline
EfficientNetB0 \cite{tan2019efficientnet} & 46\% & \textbf{91\%} & 90\% \\ \hline
EfficientNetB5 \cite{tan2019efficientnet} & 48\% & \textbf{91\%} & 0.89.6 \\ \hline
\end{tabular}
\end{table}

\begin{table}[]
\centering
\caption{Regression results of IPF lung disease progression using various regression models and MARL V1 \& V2 on different data modalities.}\label{reg_mm}
\begin{tabular}{|p{2cm}|p{0.5cm}|p{0.5cm}|p{0.5cm}|c|r|}
\hline
Model & Image & Bio & ViS & Bio+ViS & Multimodal \\ \hline
Linear & 28\% & 50\% & 10\% & 50\% & 83\% \\ \hline
LASSO & 28\% & 58\% & 10\% & 59.87\% & 82\% \\ \hline
Decision Tree & 10\% & 23.80\% & 6\% & 24.02\% & 81\% \\ \hline
Ridge & 27\% & 57\% & 7\% & 57\% & 82.90\% \\ \hline
ElasticNet & 16\% & 35\% & 18\% & 67\% & 64\% \\ \hline
Neural Network & 15\% & 32\% & 12\% & 44\% & 67\% \\ \hline
MARL V1 & \_ & \_ & \_ & \_ & \textbf{91\%} \\ \hline
MARL V2 & \_ & \_ & \_ & \_ & 89.60\% \\ \hline
\end{tabular}
\end{table}

\subsection{IPF Disease Status Classification}
\begin{figure*}
    \centering
\includegraphics[width=1.0\textwidth]{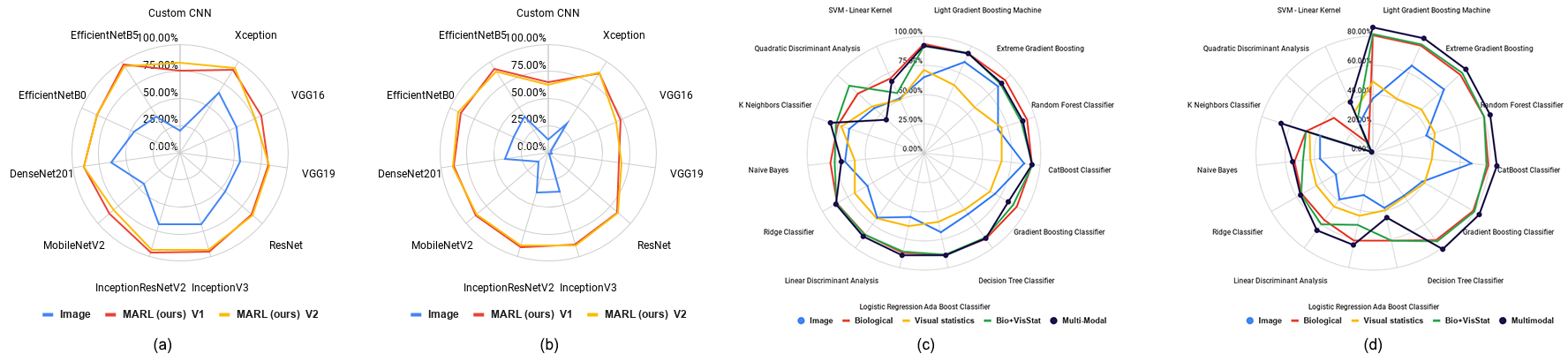}
    \caption{
Radar-charts visualising the performances of multi-class classification models using different data modalities.}
    \label{radar-MARL-binary-mc-CNN-mm}
\end{figure*}

We evaluate the proposed MARL on binary and multi-class classification tasks. For the binary discritisation, we categorise data instances based on if the score of $FVC >=2500$. For multi-class categorisation, the percent column is utilised to have three different classes, namely, sever (up to $60\%$), mild ($60\%$ to $80\%$), and good (above $80\%$).
Table \ref{binary_multi} lists the performance results of the binary and multi-class IPF lung disease status classification. Consistent with the regression results, MARL V1 and V2 improve the classification performance of state-of-the-art CNN models. The accuracy improvements range from $19\%$ to $57\%$ as shown in Table \ref{reg} and Fig. \ref{radar-MARL-binary-mc-CNN-mm} (a) and (b). Besides, we evaluate the utilisation of each data modality and their combinations as in the regression scenarios, see Table \ref{binary_multi_mm} and Fig. \ref{radar-MARL-binary-mc-CNN-mm} (c) and (d). Using the biological dataset has better results in the binary tasks than the multi-class where adding the visual statistical features contributes to higher performances. Models that use multiple data sources keep having the best results especially with our proposed MARL that has $97\%$ and $92\%$ accuracy for binary and multi-class classification, respectively.

\begin{table}[]
\centering
\caption{Results of binary and multi-class classification for IPF lung disease progression.}\label{binary_multi}
\begin{tabular}{|p{2cm}|p{0.5cm}|p{0.6cm}|p{0.6cm}|p{0.5cm}|p{0.6cm}|p{0.6cm}|}
\hline
\multicolumn{1}{|c|}{\multirow{2}{*}{Model}} & \multicolumn{3}{c|}{Binary} & \multicolumn{3}{c|}{Multi-Class} \\ \cline{2-7} 
& Image & MARL V1 & MARL V2 & Image & MARL V1 & 
MARL V2 \\ \hline
Custom CNN & 20\% & 76\% & 83\% & 12\% & 65\% & 63\% \\ \hline
VGG19 \cite{simonyan2014very} & 56\% & 82\% & 83\% & 3\% & 66\% & 68\% \\ \hline
VGG16 \cite{simonyan2014very} & 57\% & 82\% & 76\% & 3\% & 73\% & 69\% \\ \hline
ResNet \cite{he2016deep} & 55\% & 87\% & 88\% & 1.20\% & 84\% & 85\% \\ \hline
Mob.NetV2 \cite{Sandler_2018_CVPR} & 44\% & 86\% & 81\% & 12\% & 88\% & 87\% \\ \hline
Xception \cite{chollet2017xception} & 66\% & 91\% & 93\% & 32\% & 87\% & 88\% \\ \hline
Inc.ResV2 \cite{szegedy2017inception} & 69\% & 96\% & 93.50\% & 38\% & 91\% & 88.9\% \\ \hline
InceptionV3 \cite{szegedy2016rethinking} & 69\% & 95\% & 93\% & 37\% & 88\% & 89\% \\ \hline
Dens.N.201 \cite{huang2017densely} & 64\% & 90\% & 89.5\% & 40\% & 88\% & 89\% \\ \hline
Eff.tNetB0 \cite{tan2019efficientnet} & 47\% & 84\% & 84\% & 35\% & 89\% & 91\% \\ \hline
Eff.NetB5 \cite{tan2019efficientnet} & 40\% & \textbf{97\%} & 95.30\% & 40\% & \textbf{92\%} & 89.5\% \\ \hline
\end{tabular}
\end{table}

\begin{table*}[]
\centering
\caption{Results of binary and multi-class classification for lung disease progression.}\label{binary_multi_mm}
\begin{tabular}{|l|r|c|c|c|r|r|r|c|c|c|}
\hline
\multicolumn{1}{|c|}{\multirow{2}{*}{Model}} & \multicolumn{5}{c|}{Binary Classification} & \multicolumn{5}{c|}{Multi-Class Classification} \\ \cline{2-11}  & Image & 
Bio & 
ViS & 
Bio+ViS & 
Multimodal & 
Image & 
Bio & 
ViS & 
Bio+ViS & 
Multimodal \\ \hline
Extra Trees \cite{geurts2006extremely} & 65\% & 93.35\% & 71\% & 91.23\% & 92\% & 37.34\% & 80.36\% & 49\% & 81.31\% & 85.79\% \\ \hline
Light Gradient Boosting \cite{hastie2009elements} & 85.30\% & 93.27\% & 63.50\% & 93.50\% & 93\% & 65.18\% & 80.35\% & 40.50\% & 81.20\% & 85.70\% \\ \hline
Extreme Gradient Boosting \cite{chen2016xgboost} & 85.10\% & 93.18\% & 58.33\% & 88\% & 89\% & 65.27\% & 80.26\% & 44.83\% & 82\% & 85.42\% \\ \hline
Random Forest Classifier & 66.28\% & 92.90\% & 70\% & 87\% & 88.70\% & 38.53\% & 79.88\% & 44.67\% & 79.94\% & 84\% \\ \hline
Random Forest \cite{breiman2001random} & 86\% & 92.90\% & 66.67\% & 93.40\% & 93\% & 67.41\% & 79.33\% & 40.50\% & 78\% & 85\% \\ \hline
CatBoost \cite{prokhorenkova2017catboost} & 70\% & 91.51\% & 65.17\% & 88\% & 83.20\% & 38.99\% & 79.05\% & 40.50\% & 80.02\% & 84\% \\ \hline
Gradient Boosting \cite{chen2016xgboost} & 64\% & 90.32\% & 59.50\% & 89\% & 90\% & 35.74\% & 73.44\% & 36.67\% & 74.50\% & 81\% \\ \hline
Decision Tree \cite{hastie2009elements} & 69\% & 88.38\% & 60\% & 88.50\% & 89\% & 38.25\% & 60.88\% & 39.67\% & 61\% & 45\% \\ \hline
Ada Boost \cite{hastie2009multi} & 55.90\% & 87.08\% & 63.50\% & 85.83\% & 89\% & 29.43\% & 60.86\% & 43.67\% & 50.17\% & 64\% \\ \hline
Logistic Regression \cite{defazio2014saga} & 68\% & 85.97\% & 68.50\% & 86.10\% & 88\% & 39.08\% & 56.35\% & 45.17\% & 60\% & 65\% \\ \hline
Ridge Classifier & 55.90\% & 85.79\% & 68.50\% & 86.23\% & 87\% & 29.43\% & 55.90\% & 44.33\% & 56.67\% & 57\% \\ \hline
Naive Bayes & 68\% & 80.06\% & 59.67\% & 76.50\% & 71\% & 36.39\% & 53.78\% & 43.17\% & 48.17\% & 55\% \\ \hline
K Neighbors Classifier & 67\% & 77.67\% & 74\% & 78.83\% & 84\% & 37.98\% & 47.88\% & 45.50\% & 48\% & 66\% \\ \hline
SVM - Linear Kernel & 51\% & 69.58\% & 49.67\% & 56\% & 67\% & 22.20\% & 7.11\% & 30.67\% & 25.50\% & 38\% \\ \hline
Our V1 & \_ & \_ & \_ & \_ & \textbf{97\%} & \_ & \_ & \_ & \_ & \textbf{92\%} \\ \hline
Our V2 & \_ & \_ & \_ & \_ & 95.30\% & \_ & \_ & \_ & \_ & 89.50\% \\ \hline
\end{tabular}
\end{table*}





\section{Conclusion}\label{con}
We presented a novel multimodal attentional neural network architecture for representation learning. MARL, the proposed model, significantly improve accuracy to regress and classify IPF lung disease progression over state-of-the-art models. Multimodal data enable learning better feature representations than single sources. MARL includes several components designed to overcome uncertainties in the lung disease prediction when performing lung segmentation and feature representation learning. It is worthy to generalise the proposed architecture on other and different applications.

\section*{Acknowledgement}
Ali Hamdi is supported by RMIT Research Stipend Scholarship. 

\bibliographystyle{IEEEtran}
\bibliography{MARL}

\begin{thebibliography}{10}
\providecommand{\url}[1]{#1}
\csname url@samestyle\endcsname
\providecommand{\newblock}{\relax}
\providecommand{\bibinfo}[2]{#2}
\providecommand{\BIBentrySTDinterwordspacing}{\spaceskip=0pt\relax}
\providecommand{\BIBentryALTinterwordstretchfactor}{4}
\providecommand{\BIBentryALTinterwordspacing}{\spaceskip=\fontdimen2\font plus
\BIBentryALTinterwordstretchfactor\fontdimen3\font minus
  \fontdimen4\font\relax}
\providecommand{\BIBforeignlanguage}[2]{{%
\expandafter\ifx\csname l@#1\endcsname\relax
\typeout{** WARNING: IEEEtran.bst: No hyphenation pattern has been}%
\typeout{** loaded for the language `#1'. Using the pattern for}%
\typeout{** the default language instead.}%
\else
\language=\csname l@#1\endcsname
\fi
#2}}
\providecommand{\BIBdecl}{\relax}
\BIBdecl

\bibitem{cottin2014diagnosis}
V.~Cottin, B.~Crestani, D.~Valeyre, B.~Wallaert, J.~Cadranel, J.-C. Dalphin,
  P.~Delaval, D.~Israel-Biet, R.~Kessler, M.~Reynaud-Gaubert \emph{et~al.},
  ``Diagnosis and management of idiopathic pulmonary fibrosis: French practical
  guidelines,'' \emph{European Respiratory Review}, vol.~23, no. 132, pp.
  193--214, 2014.

\bibitem{kafaja2018reliability}
S.~Kafaja, P.~J. Clements, H.~Wilhalme, C.-h. Tseng, D.~E. Furst, G.~H. Kim,
  J.~Goldin, E.~R. Volkmann, M.~D. Roth, D.~P. Tashkin \emph{et~al.},
  ``Reliability and minimal clinically important differences of fvc. results
  from the scleroderma lung studies (sls-i and sls-ii),'' \emph{American
  journal of respiratory and critical care medicine}, vol. 197, no.~5, pp.
  644--652, 2018.

\bibitem{hamdi2021spatiotemporal}
A.~Hamdi, K.~Shaban, A.~Erradi, A.~Mohamed, S.~K. Rumi, and F.~D. Salim,
  ``Spatiotemporal data mining: a survey on challenges and open problems,''
  \emph{Artificial Intelligence Review}, pp. 1--48, 2021.

\bibitem{tripathy2014image}
B.~Tripathy, A.~Basu, and S.~Govel, ``Image segmentation using spatial
  intuitionistic fuzzy c means clustering,'' in \emph{2014 IEEE International
  Conference on Computational Intelligence and Computing Research}.\hskip 1em
  plus 0.5em minus 0.4em\relax IEEE, 2014, pp. 1--5.

\bibitem{tan2019efficientnet}
M.~Tan and Q.~Le, ``Efficientnet: Rethinking model scaling for convolutional
  neural networks,'' in \emph{International Conference on Machine
  Learning}.\hskip 1em plus 0.5em minus 0.4em\relax PMLR, 2019, pp. 6105--6114.

\bibitem{luo2016understanding}
W.~Luo, Y.~Li, R.~Urtasun, and R.~Zemel, ``Understanding the effective
  receptive field in deep convolutional neural networks,'' in \emph{Advances in
  neural information processing systems}, 2016, pp. 4898--4906.

\bibitem{IPF_kaggle_2020}
\BIBentryALTinterwordspacing
O.~S. I.~C. (OSIC), 2020. [Online]. Available:
  \url{https://www.kaggle.com/c/osic-pulmonary-fibrosis-progression/}
\BIBentrySTDinterwordspacing

\bibitem{zhang2019aet}
L.~Zhang, G.-J. Qi, L.~Wang, and J.~Luo, ``Aet vs. aed: Unsupervised
  representation learning by auto-encoding transformations rather than data,''
  in \emph{Proceedings of the IEEE Conference on Computer Vision and Pattern
  Recognition}, 2019, pp. 2547--2555.

\bibitem{kolesnikov2019revisiting}
A.~Kolesnikov, X.~Zhai, and L.~Beyer, ``Revisiting self-supervised visual
  representation learning,'' in \emph{Proceedings of the IEEE conference on
  Computer Vision and Pattern Recognition}, 2019, pp. 1920--1929.

\bibitem{hamdi2020drotrack}
A.~Hamdi, F.~Salim, and D.~Y. Kim, ``Drotrack: High-speed drone-based object
  tracking under uncertainty,'' in \emph{Proceedings of the IEEE conference on
  Fuzzy Systems (FUZZ-IEEE)}, 2020.

\bibitem{yamashita2018convolutional}
R.~Yamashita, M.~Nishio, R.~K.~G. Do, and K.~Togashi, ``Convolutional neural
  networks: an overview and application in radiology,'' \emph{Insights into
  imaging}, vol.~9, no.~4, pp. 611--629, 2018.

\bibitem{raghu2011official}
G.~Raghu, H.~R. Collard, J.~J. Egan, F.~J. Martinez, J.~Behr, K.~K. Brown,
  T.~V. Colby, J.-F. Cordier, K.~R. Flaherty, J.~A. Lasky \emph{et~al.}, ``An
  official ats/ers/jrs/alat statement: idiopathic pulmonary fibrosis:
  evidence-based guidelines for diagnosis and management,'' \emph{American
  journal of respiratory and critical care medicine}, vol. 183, no.~6, pp.
  788--824, 2011.

\bibitem{bueno2018updated}
J.~Bueno, L.~Landeras, and J.~H. Chung, ``Updated fleischner society guidelines
  for managing incidental pulmonary nodules: common questions and challenging
  scenarios,'' \emph{Radiographics}, vol.~38, no.~5, pp. 1337--1350, 2018.

\bibitem{walsh2018deep}
S.~L. Walsh, L.~Calandriello, M.~Silva, and N.~Sverzellati, ``Deep learning for
  classifying fibrotic lung disease on high-resolution computed tomography: a
  case-cohort study,'' \emph{The Lancet Respiratory Medicine}, vol.~6, no.~11,
  pp. 837--845, 2018.

\bibitem{simonyan2014very}
K.~Simonyan and A.~Zisserman, ``Very deep convolutional networks for
  large-scale image recognition,'' \emph{arXiv preprint arXiv:1409.1556}, 2014.

\bibitem{he2016deep}
K.~He, X.~Zhang, S.~Ren, and J.~Sun, ``Deep residual learning for image
  recognition,'' in \emph{Proceedings of the IEEE conference on computer vision
  and pattern recognition}, 2016, pp. 770--778.

\bibitem{huang2016deep}
G.~Huang, Y.~Sun, Z.~Liu, D.~Sedra, and K.~Q. Weinberger, ``Deep networks with
  stochastic depth,'' in \emph{European conference on computer vision}.\hskip
  1em plus 0.5em minus 0.4em\relax Springer, 2016, pp. 646--661.

\bibitem{he2016identity}
K.~He, X.~Zhang, S.~Ren, and J.~Sun, ``Identity mappings in deep residual
  networks,'' in \emph{European conference on computer vision}.\hskip 1em plus
  0.5em minus 0.4em\relax Springer, 2016, pp. 630--645.

\bibitem{huang2017densely}
G.~Huang, Z.~Liu, L.~Van Der~Maaten, and K.~Q. Weinberger, ``Densely connected
  convolutional networks,'' in \emph{Proceedings of the IEEE conference on
  computer vision and pattern recognition}, 2017, pp. 4700--4708.

\bibitem{Sandler_2018_CVPR}
M.~Sandler, A.~Howard, M.~Zhu, A.~Zhmoginov, and L.-C. Chen, ``Mobilenetv2:
  Inverted residuals and linear bottlenecks,'' in \emph{The IEEE Conference on
  Computer Vision and Pattern Recognition (CVPR)}, June 2018.

\bibitem{howard2017mobilenets}
A.~G. Howard, M.~Zhu, B.~Chen, D.~Kalenichenko, W.~Wang, T.~Weyand,
  M.~Andreetto, and H.~Adam, ``Mobilenets: Efficient convolutional neural
  networks for mobile vision applications,'' \emph{arXiv preprint
  arXiv:1704.04861}, 2017.

\bibitem{zoph2018learning}
B.~Zoph, V.~Vasudevan, J.~Shlens, and Q.~V. Le, ``Learning transferable
  architectures for scalable image recognition,'' in \emph{Proceedings of the
  IEEE conference on computer vision and pattern recognition}, 2018, pp.
  8697--8710.

\bibitem{chollet2017xception}
F.~Chollet, ``Xception: Deep learning with depthwise separable convolutions,''
  in \emph{Proceedings of the IEEE conference on computer vision and pattern
  recognition}, 2017, pp. 1251--1258.

\bibitem{hamdi2021flexgrid2vec}
A.~Hamdi, D.~Y. Kim, and F.~Salim, ``flexgrid2vec: Learning efficient visual
  representations vectors,'' \emph{arXiv e-prints}, pp. arXiv--2007, 2021.

\bibitem{masi2016we}
I.~Masi, A.~T. Tran, T.~Hassner, J.~T. Leksut, and G.~Medioni, ``Do we really
  need to collect millions of faces for effective face recognition?'' in
  \emph{European Conference on Computer Vision}.\hskip 1em plus 0.5em minus
  0.4em\relax Springer, 2016, pp. 579--596.

\bibitem{hong2019predicting}
X.~Hong, R.~Lin, C.~Yang, N.~Zeng, C.~Cai, J.~Gou, and J.~Yang, ``Predicting
  alzheimer’s disease using lstm,'' \emph{IEEE Access}, vol.~7, pp.
  80\,893--80\,901, 2019.

\bibitem{gu2019method}
J.~Gu, L.~Liang, H.~Song, Y.~Kong, R.~Ma, Y.~Hou, J.~Zhao, J.~Liu, N.~He, and
  Y.~Zhang, ``A method for hand-foot-mouth disease prediction using geodetector
  and lstm model in guangxi, china,'' \emph{Scientific reports}, vol.~9, no.~1,
  pp. 1--10, 2019.

\bibitem{chimmula2020time}
V.~K.~R. Chimmula and L.~Zhang, ``Time series forecasting of covid-19
  transmission in canada using lstm networks,'' \emph{Chaos, Solitons \&
  Fractals}, vol. 135, p. 109864, 2020.

\bibitem{gao2019distanced}
R.~Gao, Y.~Huo, S.~Bao, Y.~Tang, S.~L. Antic, E.~S. Epstein, A.~B. Balar,
  S.~Deppen, A.~B. Paulson, K.~L. Sandler \emph{et~al.}, ``Distanced lstm:
  time-distanced gates in long short-term memory models for lung cancer
  detection,'' in \emph{International Workshop on Machine Learning in Medical
  Imaging}.\hskip 1em plus 0.5em minus 0.4em\relax Springer, 2019, pp.
  310--318.

\bibitem{marentakis2021lung}
P.~Marentakis, P.~Karaiskos, V.~Kouloulias, N.~Kelekis, S.~Argentos,
  N.~Oikonomopoulos, and C.~Loukas, ``Lung cancer histology classification from
  ct images based on radiomics and deep learning models,'' \emph{Medical \&
  Biological Engineering \& Computing}, pp. 1--12, 2021.

\bibitem{tang2020elaboration}
X.~Tang, X.~Xu, Z.~Han, G.~Bai, H.~Wang, Y.~Liu, P.~Du, Z.~Liang, J.~Zhang,
  H.~Lu \emph{et~al.}, ``Elaboration of a multimodal mri-based radiomics
  signature for the preoperative prediction of the histological subtype in
  patients with non-small-cell lung cancer,'' \emph{Biomedical engineering
  online}, vol.~19, no.~1, p.~5, 2020.

\bibitem{li2020deep}
L.~Li, X.~Zhao, W.~Lu, and S.~Tan, ``Deep learning for variational
  multimodality tumor segmentation in pet/ct,'' \emph{Neurocomputing}, vol.
  392, pp. 277--295, 2020.

\bibitem{subramanian2020multimodal}
V.~Subramanian, M.~N. Do, and T.~Syeda-Mahmood, ``Multimodal fusion of imaging
  and genomics for lung cancer recurrence prediction,'' in \emph{2020 IEEE 17th
  International Symposium on Biomedical Imaging (ISBI)}.\hskip 1em plus 0.5em
  minus 0.4em\relax IEEE, 2020, pp. 804--808.

\bibitem{qiao2019mnn}
Z.~Qiao, X.~Wu, S.~Ge, and W.~Fan, ``Mnn: multimodal attentional neural
  networks for diagnosis prediction,'' \emph{Extraction}, vol.~1, p.~A1, 2019.

\bibitem{zagoruyko2016wide}
S.~Zagoruyko and N.~Komodakis, ``Wide residual networks,'' in \emph{British
  Machine Vision Conference 2016}.\hskip 1em plus 0.5em minus 0.4em\relax
  British Machine Vision Association, 2016.

\bibitem{ioffe2015batch}
S.~Ioffe and C.~Szegedy, ``Batch normalization: Accelerating deep network
  training by reducing internal covariate shift,'' in \emph{International
  conference on machine learning}.\hskip 1em plus 0.5em minus 0.4em\relax PMLR,
  2015, pp. 448--456.

\bibitem{tan2019mnasnet}
M.~Tan, B.~Chen, R.~Pang, V.~Vasudevan, M.~Sandler, A.~Howard, and Q.~V. Le,
  ``Mnasnet: Platform-aware neural architecture search for mobile,'' in
  \emph{Proceedings of the IEEE/CVF Conference on Computer Vision and Pattern
  Recognition}, 2019, pp. 2820--2828.

\bibitem{luong2015effective}
M.-T. Luong, H.~Pham, and C.~D. Manning, ``Effective approaches to
  attention-based neural machine translation,'' in \emph{Proceedings of the
  2015 Conference on Empirical Methods in Natural Language Processing}, 2015,
  pp. 1412--1421.

\bibitem{szegedy2016rethinking}
C.~Szegedy, V.~Vanhoucke, S.~Ioffe, J.~Shlens, and Z.~Wojna, ``Rethinking the
  inception architecture for computer vision,'' in \emph{Proceedings of the
  IEEE conference on computer vision and pattern recognition}, 2016, pp.
  2818--2826.

\bibitem{szegedy2017inception}
C.~Szegedy, S.~Ioffe, V.~Vanhoucke, and A.~Alemi, ``Inception-v4,
  inception-resnet and the impact of residual connections on learning,'' in
  \emph{Proceedings of the AAAI Conference on Artificial Intelligence},
  vol.~31, no.~1, 2017.

\bibitem{geurts2006extremely}
P.~Geurts, D.~Ernst, and L.~Wehenkel, ``Extremely randomized trees,''
  \emph{Machine learning}, vol.~63, no.~1, pp. 3--42, 2006.

\bibitem{hastie2009elements}
T.~Hastie, R.~Tibshirani, and J.~Friedman, \emph{The elements of statistical
  learning: data mining, inference, and prediction}.\hskip 1em plus 0.5em minus
  0.4em\relax Springer Science \& Business Media, 2009.

\bibitem{chen2016xgboost}
T.~Chen and C.~Guestrin, ``Xgboost: A scalable tree boosting system,'' in
  \emph{Proceedings of the 22nd acm sigkdd international conference on
  knowledge discovery and data mining}, 2016, pp. 785--794.

\bibitem{breiman2001random}
L.~Breiman, ``Random forests,'' \emph{Machine learning}, vol.~45, no.~1, pp.
  5--32, 2001.

\bibitem{prokhorenkova2017catboost}
L.~Prokhorenkova, G.~Gusev, A.~Vorobev, A.~V. Dorogush, and A.~Gulin,
  ``Catboost: unbiased boosting with categorical features,'' \emph{arXiv
  preprint arXiv:1706.09516}, 2017.

\bibitem{hastie2009multi}
T.~Hastie, S.~Rosset, J.~Zhu, and H.~Zou, ``Multi-class adaboost,''
  \emph{Statistics and its Interface}, vol.~2, no.~3, pp. 349--360, 2009.

\bibitem{defazio2014saga}
A.~Defazio, F.~R. Bach, and S.~Lacoste-Julien, ``Saga: A fast incremental
  gradient method with support for non-strongly convex composite objectives,''
  in \emph{NIPS}, 2014.

\end{thebibliography}

\end{document}